\documentclass[pmlr]{jmlr}
\RequirePackage{graphicx}
 \usepackage{booktabs}
\usepackage{longtable}
\makeatletter
\def\set@curr@file#1{\def\@curr@file{#1}} 
\makeatother
\usepackage[load-configurations=version-1]{siunitx} 

\theorembodyfont{\upshape}
\theoremheaderfont{\scshape}
\theorempostheader{:}
\theoremsep{\newline}

\usepackage{amsmath,amssymb,mathtools}

\makeatletter
\def\@titlefoot{}
\def\ps@jmlrtps{%
  \let\@mkboth\@gobbletwo
  \def\@oddhead{}%
  \let\@evenhead\@oddhead
  \def\@oddfoot{}%
  \let\@evenfoot\@oddfoot
}
\makeatother

\title[Targeted Question Selection for Psychiatric Intake]{Adaptive Question Selection from a Large Question Bank for Field Recovery in Conversational Psychiatric Intake}

\author{\Name{Shevya Panda, } \Name{Shinjini Bose} \and \Name{Ananya Joshi}\\
\addr Johns Hopkins University\\ Baltimore, USA}

\author{\Name{Guan Gui, } \Name{Peter Zandi,} \Name{Jacob Taylor,} \and \Name{Ananya Joshi}\\
\addr Johns Hopkins University\\ Baltimore, USA}

\begin{document}

\maketitle

\begin{abstract}
Psychiatric intake is a sequential, high-stakes information-gathering process in which clinicians must decide what to ask, in what order, and how to interpret incomplete or ambiguous responses under limited time. Despite growing interest in conversational AI for healthcare, there is still limited infrastructure for conversational AI in this application. Accordingly, we formulate this task as a question-selection problem with clinically grounded questions, known target information, and controllable patient difficulty. We also introduce a task-specific question-selection benchmark based on a bank of 655 clinician-authored intake questions and corresponding synthetic patient vignettes with 5 different behavioral conditions. In our evaluation, we compare random questioning, a clinical psychiatric intake form baseline, and an LLM-guided adaptive policy across 300 interview sessions spanning four patients and five behavioral conditions. Across the benchmark, the clinically ordered fixed form substantially outperforms random questioning, and the LLM-guided policy achieves the strongest overall recovery. The advantage of adaptation grows sharply under patient behavior that is less amenable to field recovery, especially under guarded-concise conditions. These findings suggest that performance in conversational clinical systems depends not only on language understanding after information is disclosed, but also on whether the system reaches the right topics within a limited interaction budget. More broadly, the benchmark provides a controlled framework for studying how clinical structure and adaptive follow-up contribute to information recovery in interactive clinical machine learning.
\end{abstract}

\section{Introduction}
Psychiatric intake is an adaptive information-gathering process in which clinicians use open-ended and targeted follow-up questions to understand a patient’s condition \citep{silverman2015psycheval,hashim2017patientcentered}. During intake, clinicians must decide which questions to ask, in what order, and how to interpret partial or ambiguous answers to support downstream care \citep{silverman2015psycheval}. Because responses are often subjective, incomplete, and context-dependent, what to ask next depends on what the patient has already discussed \citep{hashim2017patientcentered}. Unlike other medical domains with structured measurements, these aspects make psychiatric intake an adaptive process under uncertainty, where effective care depends on selecting the most informative next question \citep{silverman2015psycheval,hashim2017patientcentered}.

Still, in practice, psychiatric assessment is often operationalized through brief fixed-form screening tools and structured questionnaires, particularly for common symptom domains such as depression and anxiety \citep{kroenke2001phq9,spitzer2006gad7,taylor2018internethistory}. Form-based approaches like these may assume that relevant questions can be specified in advance, reducing intake to a fixed sequence rather than an adaptive questioning process. However, this process can be frustrating to patients and produce incomplete or misleading representations of a patient. This is especially concerning in fragmented care systems, where the intake results may be used by downstream clinicians who lack access to the original intake context \citep{silverman2015psycheval,hashim2017patientcentered}. Therefore, we need new algorithms for question selection that optimize information gathering under time constraints and uncertainty, rather than treating intake as static data collection. This task is challenging because such systems must recover safety-, treatment-, and social-context information within a limited conversational window, often from patients who differ substantially in how much they disclose and how directly they answer \citep{silverman2015psycheval,weber2017suicidal,hallford2023disclosure}. Missing the right question can mean missing critical medical information, like current suicidal ideation, unhealthy alcohol or tobacco use, prior psychiatric treatment, or medically relevant background \citep{weber2017suicidal,jointcommission2018r3suicide,uspstf2018alcohol,uspstf2021tobacco}. 

We formalize psychiatric intake as a machine learning problem in which questions are selected sequentially from a large, fixed question bank to maximize recovery of clinically relevant information under a constrained interview budget. Through a participatory design process with clinicians, we construct a benchmark from 655 real intake questions spanning multiple psychiatric settings (e.g. substance use clinic). Within this framework, we compare three question-selection strategies that differ in how much clinical structure and adaptation they use. We pair this question bank with clinician-informed synthetic vignettes that generate responses under controlled behavioral conditions (default, forthcoming-talkative, forthcoming-concise, guarded-talkative, and guarded-concise) \citep{kononowicz2019virtual}, and evaluate the resulting interactions under a preregistered protocol. Specifically, we compare a random baseline, a clinically ordered fixed-form baseline, and an LLM-guided adaptive policy, and we review transcripts for hallucinated or unsupported information.

The evaluation yields three main findings. First, clinically structured questioning substantially outperforms untargeted random questioning across the full benchmark. Second, the advantage of adaptation grows as patient behavior becomes less amenable to field recovery, especially under guarded and concise response styles. Third, the benchmark shows that the main bottleneck in this setting is not only whether a system can interpret disclosed information, but whether it reaches the right topics within a limited conversational budget. These results suggest that interactive clinical systems should be evaluated not only by end-task accuracy, but also by how they allocate limited conversational opportunities across clinically meaningful topics.

\subsection*{Generalizable Insights about Machine Learning in the Context of Healthcare}

\begin{enumerate}
    \item In interactive clinical tasks, question selection is itself an ML task. Specific to psychiatric intake, across subdivisions, a good intake process depends not only on language understanding after information is disclosed, but also on whether the system reaches the right topics quickly.

    \item Controlled synthetic evaluation can reveal clinically relevant failure modes before deployment. By varying patient behavioral condition while preserving solvability, the benchmark exposes degradation patterns, robustness differences, and efficiency tradeoffs that would be difficult to isolate in unconstrained case studies.
\end{enumerate}

\section{Related Work}

Psychiatric intake is distinct from many other clinical data-collection settings because the relevant information is often subjective, incomplete, and only partially disclosed at the time of questioning. Effective intake therefore depends not only on what information is eventually elicited, but also on how the interviewer sequences follow-up questions, resolves ambiguity, and adapts to patient behavior over time \citep{silverman2015psycheval,weber2017suicidal,hashim2017patientcentered}. Structured psychiatric interviews such as the SCID motivate fixed, clinically grounded interview formats, but they are still open to adaptive clinical judgement \citep{first2016scid}. This makes psychiatric intake different from settings in which diagnostically relevant features are available early through structured measurements or fixed forms, and motivates treating question order itself as part of the learning problem.

Conversational AI for psychiatric intake also has different considerations than other medical conversational AI systems. Previous work has examined conversational systems in healthcare and mental health \citep{laranjo2018conversational,vaidyam2019chatbots}, diagnosis-oriented medical dialog datasets \citep{zeng2020meddialog}, virtual patients for education \citep{kononowicz2019virtual}, and adaptive information acquisition in a more general way \citep{settles2012active,vanderlinden2010elements}.  Systematic reviews have emphasized both the promise and limitations of this research: many studies remain in the early stages, narrowly skewed or evaluated without strong evidence of safety, efficacy, or generalizability \citep{laranjo2018conversational,vaidyam2019chatbots}.  In mental health specifically, prior work has focused largely on chatbots for support, screening, or treatment-adjacent interaction \citep{liu2026uptake, vaidyam2019chatbots}. However, these lines of work do not provide a method for systematic psychiatric intake question selection with a clinically grounded question source, known target information, controllable patient difficulty, and a clinically meaningful fixed-form comparator. Without these elements, it is difficult to determine whether stronger performance comes from better question choice, stronger clinical structure, easier cases, or more favorable conversational dynamics.

Our work differs in both objective and evaluation setting. Rather than studying support or therapeutic interaction, we study how different questioning policies recover clinically relevant information under a fixed conversational budget, reflective of realistic psychiatric conditions that can be meaningful from an information-theoretic lens. In particular, we make use of patient vignettes to spawn virtual patient agents to respond to clinician questions. These types of virtual patients are widely used as interactive simulations of clinical scenarios \citep{kononowicz2019virtual}; here, they provide a controlled testbed for algorithmic comparison in which patient difficulty can be varied while ground-truth intake targets remain known.


At the machine learning level, the task is most closely related to active information acquisition and adaptive assessment, where the central problem is how to select the next query or measurement under a limited budget \citep{settles2012active,vanderlinden2010elements}. However, they do not address the specific challenges of psychiatric dialogue, including incidental disclosure, heterogeneous patient behavioral condition, and the mismatch between a large clinically grounded question bank and a much smaller set of evaluation targets. Our work brings this acquisition perspective into a clinically meaningful conversational setting and makes the resulting strategy tradeoffs measurable.


\section{Approach}
\begin{figure}[t]
\centering
\includegraphics[width=\linewidth]{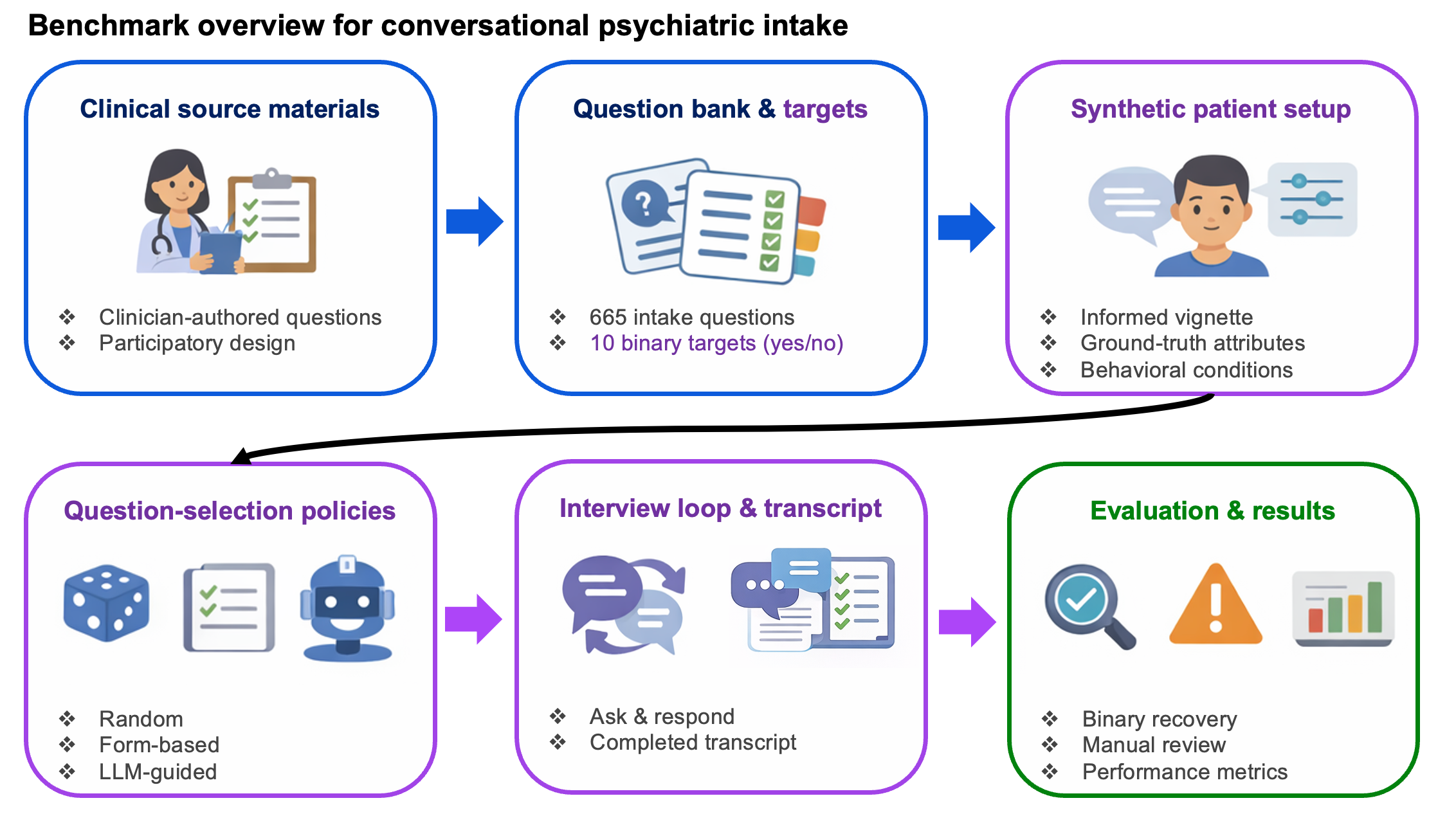}
\caption{Overview of the conversational psychiatric intake benchmark. Clinician-authored source materials and participatory design are used to construct a bank of 655 intake questions and 10 binary evaluation targets. Clinician-informed vignettes define synthetic patients with ground-truth attributes and the 5 controlled behavioral conditions. Random, form-based, and LLM-guided policies interact with these patients under a fixed interview budget, producing transcripts that are evaluated through automated binary recovery scoring and manual review.}
\label{fig:benchmark_overview}
\end{figure}

Figure~\ref{fig:benchmark_overview} summarizes the full task workflow, from question-bank construction and synthetic patient setup to interview generation, evaluation, and reported outputs.

The problem formulation and evaluation setting were informed by a participatory design process with 3 clinicians and behavioral health scientists involved in psychiatric intake across multiple care settings. These discussions characterized the structure of intake and the criteria for a high-quality clinical interaction. Clinicians described psychiatric intake as a modular process organized around key targets such as chief complaint, psychiatric history, family history, substance use, and social context. They emphasized that effective intake prioritizes safety and risk assessment, core symptom characterization, severity and functional impairment, and temporal course \citep{apa2006evaluation,first2016scid,sheehan1998mini}. They also highlighted that high-quality interviews must resolve ambiguity, detect when patients may be guarded or unreliable, distinguish between psychiatric and non-psychiatric contributors, and incorporate patient goals and treatment preferences. 

Clinicians advised restricting the system to a vetted set of real intake questions to preserve clinical validity and safety. While unconstrained language models may elicit more information in some cases, they may also introduce unsafe, inappropriate, or clinically misleading queries. This constraint defines the action space of our methods as selection over a fixed, clinically grounded question bank. They also identified the importance of the behavioral condition of the patient during the intake.  A question that is effective for a forthcoming patient may be far less useful for a guarded or highly concise patient, especially when disclosure requires direct prompting. A high-quality psychiatric interview was generally defined as one that, among other aspects, identifies safety concerns, clarifies ambiguity, assesses symptom severity and functional impact, gathers information necessary for diagnosis and treatment planning, detects unreliable reporting, distinguishes psychiatric from medical or substance-related contributors, and incorporates patient goals and preferences. These criteria guided the selection of target fields and evaluation metrics used in our experiments \citep{apa2006evaluation}.

Accordingly, we propose question selection for psychiatric intake as a machine learning problem. Under a fixed budget of conversational turns, the central challenge is not only to interpret what the patient says, but also to decide what to ask next. Framed this way, intake becomes a budgeted information-recovery task in which a questioning policy must allocate limited conversational turns across a large bank of clinically grounded candidate questions. This framing is related to active information acquisition and adaptive assessment \citep{settles2012active,vanderlinden2010elements}. The bank of questions used was derived from clinician-authored intake materials collected from a large Department of Psychiatry. The source material included structured intake questions used in routine workflows, form-field artifacts, and non-conversational fragments. After removing irrelevant, redundant, or non-actionable items, the final bank contains \(N=655\) questions, down from 812 raw items, spanning psychiatric history, substance use, medical history, family history, social context, and risk assessment \footnote{The raw question bank order was not treated as a clinically meaningful interview sequence: it reflected source and spreadsheet parsing order rather than a reproducible psychiatric assessment protocol. This distinction is important because the source forms themselves were designed for written pre-visit completion, not conversational assessment under a fixed 20-turn budget.}. 

\subsection{Problem Formulation}

We formalize psychiatric intake as a budgeted information-recovery problem. Let \(Q\) denote the bank of \(N=655\) candidate intake questions derived from real clinical intake practice, and let \(F=\{f_1,\dots,f_K\}\) denote a set of \(K=10\) binary target fields to be recovered from the interview. Each interaction is constrained by a fixed budget of \(T=20\) question turns. At turn \(t\), a question-selection policy \(\pi\) observes the dialogue history, which is a question and response pair, \(H_t = \{(q_1, r_1), \dots, (q_{t-1}, r_{t-1})\}\) and selects the next question \(q_t \sim \pi(\cdot \mid H_t)\). Patient responses are generated by an external LLM conditioned on the patient vignette, behavioral condition, and prior dialogue context. Response length and conversational detail may still vary across interviews, but the primary benchmark constraint is the fixed number of turns. The objective is therefore to maximize expected recovery accuracy under this interaction budget:

\begin{equation}
\label{eq:objective}
\max_{\pi \in \Pi_T}\ \mathbb{E}\!\left[\frac{1}{K}\sum_{k=1}^{K}
\mathbb{I}\!\left({\pi}(f_k)=f_k\right)\right]
\end{equation}

where this expectation is taken over the policy and synthetic patient behavior. We approximate this value empirically by evaluating each policy across benchmark patients, behavioral conditions, and repeated runs, then averaging transcript-level recovery accuracy. After each interview, the transcript is evaluated against the ground-truth evaluation target.



\paragraph{Data Provenance}
The question bank was derived from real clinician question data collected from psychiatric intake practice. No real patient transcript data were used in the benchmark experiments. All evaluated conversations were conducted with authored synthetic patient profiles created by the research team and not derived from patient records. 


\subsection{Synthetic Patient Simulation and Controlled Behavioral Variation}

We evaluate question-selection policies using authored synthetic psychiatric patients rather than real patient transcripts. Patient behavior is simulated using GPT-4o-family models \citep{openai2024gpt4o}, conditioned on structured vignettes. This setup enables controlled variation in patient characteristics while preserving known ground-truth information.

Each synthetic patient profile includes a vignette, structured background context, a specified communication style, and ground-truth values for all 10 binary evaluation fields. Four evaluation patients were selected from a pool of 12 authored profiles to maximize variation in diagnostic presentation, severity, native communication style, and ground-truth field distribution. The final set spans major depressive disorder (MDD), generalized anxiety disorder (GAD), and post-traumatic stress disorder (PTSD), with one severe MDD case that also includes current suicidal ideation. The four evaluation patients are labeled Eval001--Eval004 in the reported benchmark results.

To study the effect of patient behavior on information recovery, we vary patient behavior along two axes: cooperativeness (forthcoming vs.\ guarded) and verbosity (talkative vs.\ concise). Forthcoming patients typically answer direct questions on the first attempt, whereas guarded patients may defer or partially answer before yielding on follow-up. Talkative patients elaborate and may volunteer related detail, whereas concise patients respond briefly and rarely expand beyond the question asked. These settings are instantiated over the same underlying vignette and ground-truth attributes, creating multiple interaction variants with controlled difficulty. A default condition preserves each patient’s native communication style without modification. The cooperativeness and verbosity manipulations should therefore be interpreted as a structured difficulty gradient rather than as fully independent behavioral factors. To monitor simulation fidelity, we perform manual transcript review for inconsistencies and hallucinations. 


\begin{figure}[t]
\centering
\includegraphics[width=\linewidth]{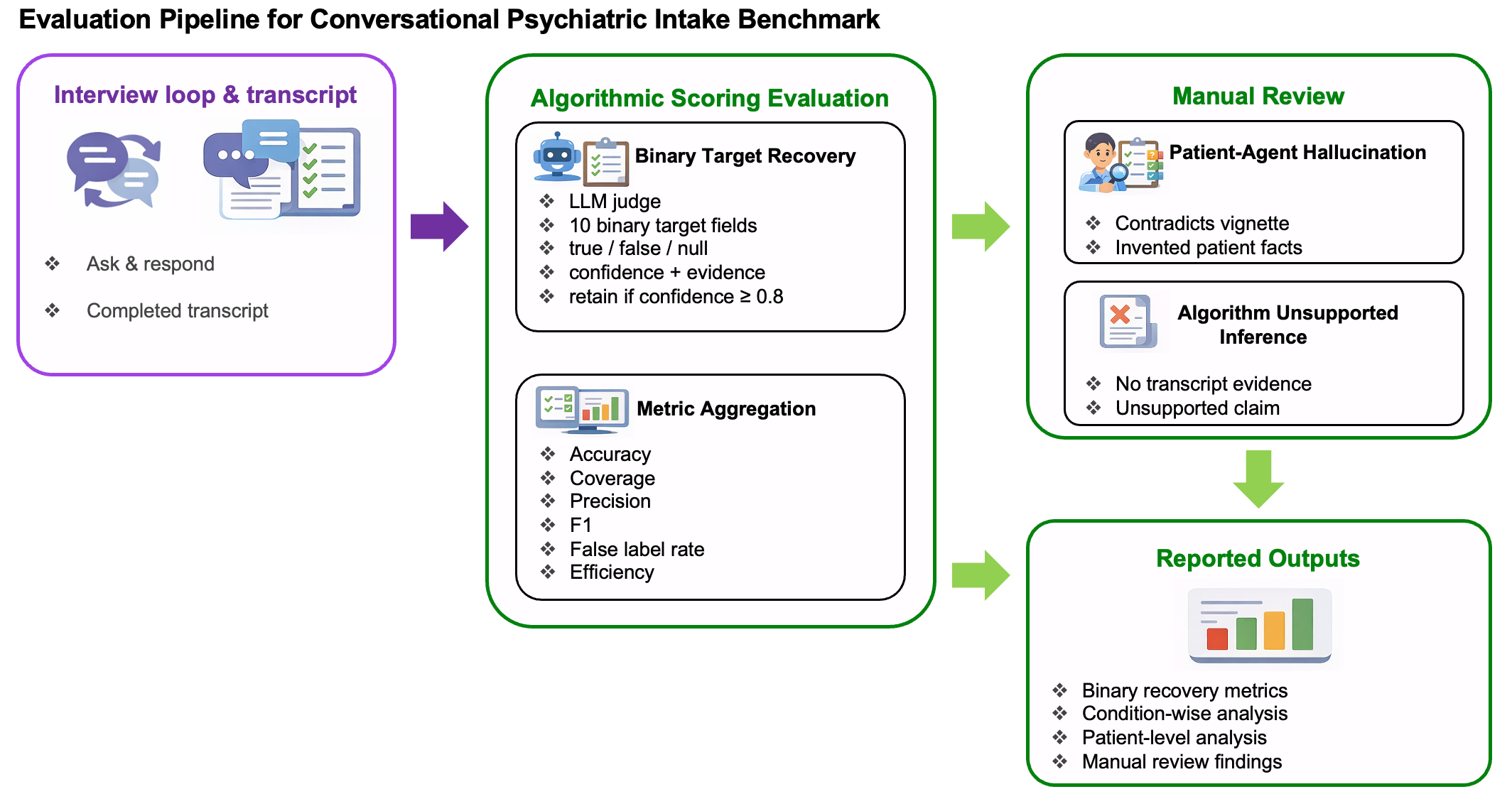}
\caption{Evaluation pipeline for the conversational psychiatric intake benchmark. Each completed transcript is evaluated in two stages. Automated binary evaluation uses a post-session LLM judge to score recovery of the 10 binary target fields and aggregate transcript-level metrics, while manual review identifies patient-agent hallucinations and algorithm unsupported inferences. Reported outputs include binary recovery metrics together with condition-wise, patient-level, and manual-review summaries.}
\label{fig:eval_pipeline}
\end{figure}

\subsection{Question-Selection Policies}
We evaluated three policy classes. Across all strategies, previously asked questions are excluded from future selection within the same interview, so each policy operates without repeating exact question text.

\paragraph{Random conversational baseline.}
The random baseline samples an unanswered question uniformly at random from the bank at each turn and asks it in its original bank wording. This baseline represents an interviewer with no structure and no adaptation, using the full clinical bank without strategic targeting. Random selection serves as a lower-bound comparator for sequential information acquisition under a fixed budget \citep{settles2012active}.

\paragraph{Form-based baseline.}
The form-based baseline asks questions in a fixed clinically ordered sequence derived from published psychiatric evaluation frameworks and clinician review. Domain priority follows the systematic assessment structure described in the APA Practice Guideline for the Psychiatric Evaluation of Adults, while the modular organization is informed by structured psychiatric interviews such as the SCID and MINI \citep{apa2006evaluation,first2016scid,sheehan1998mini}. Safety screening is prioritized early in the sequence following suicide-risk assessment literature \citep{posner2011cssrs}. This ordering is different from the native order of the source intake forms, as these were designed for written pre-visit completion rather than conversational assessment under time/conversation restrictions. The ordering is constructed using domain priority, per-domain budget caps, subdomain caps, variable-key grouping to avoid redundant screening questions, and a binary-before-open ordering within most modules.

\paragraph{LLM-guided policy.}
The LLM-guided policy performs question selection online. At each turn, it samples up to 40 unanswered candidate questions uniformly at random from the bank, then presents those candidates to GPT-4o together with the conversation history and the current list of unrecovered fields. The model is prompted to select exactly one candidate question judged most likely to recover new information. Unlike fixed-form questioning, this strategy can adapt to unexpected disclosures, partial answers, and local topic shifts during the interview.

\subsection{Evaluation Pipeline}

Each completed interaction is evaluated in two stages: algorithmic scoring and manual review. Algorithmic scoring measures how much clinically relevant information can be recovered from the transcript under a fixed evaluation protocol. Manual review identifies process failures that are not fully captured by automated metrics, including hallucinations by the patient agent and unsupported inferences by the questioning algorithm. Figure~\ref{fig:eval_pipeline} summarizes this pipeline.

\paragraph{Algorithmic Scoring Evaluation.}
Given a completed transcript, a post-session LLM judge implemented with GPT-4o evaluates each of the \(K=10\) binary target fields independently. For each field, the judge returns a binary mentioned flag, an extracted value in \(\{\texttt{true}, \texttt{false}, \texttt{null}\}\), a confidence score, and an evidence quote from the transcript. Extractions with confidence below \(\tau=0.8\) are discarded to avoid hallucinated text. During the interview itself, the system maintains a lightweight in-loop field-tracking heuristic based on conservative keyword matching, but final benchmark scoring relies only on the post-session judge. This design allows the evaluation to credit both direct answers and incidental mentions that emerge during natural conversation, while discarding low-confidence extractions from the final benchmark score.

\paragraph{Manual review.}
After algorithmic scoring, each generated transcript is manually reviewed for two classes of process failures. The first is \emph{patient-agent hallucination}, in which the synthetic patient contradicts or invents information that is not supported by the underlying vignette. The second is \emph{algorithm hallucination}, in which the questioning policy or downstream answer-generation step attributes information to the patient that does not appear in the transcript. This review therefore distinguishes failures of patient simulation from failures of elicitation, extraction, or inference. We report these findings separately from the main benchmark metrics because they characterize process-level failures rather than only end-task performance.

\subsection{Metrics}
Let \(M\) denote the number of non-null extracted binary fields, \(C\) the number of correct extractions, \(L\) the number of incorrect extractions, and \(W\) the cumulative transcript word count. We report the following transcript-level metrics:
\begin{subequations}
\label{eq:metrics}
\begin{align}
\mathrm{Accuracy} &= \frac{1}{K}\sum_{k=1}^{K}
\mathbb{I}\!\left({\pi}(f_k)=f_k\right), &
\mathrm{Coverage} &= \frac{M}{K}, \\
\mathrm{Precision} &=
\begin{cases}
\frac{C}{M}, & M>0,\\
1, & M=0,
\end{cases}
&
\mathrm{F1\ score} &=
\begin{cases}
\frac{2\,\mathrm{Prec}\cdot \mathrm{Cov}}{\mathrm{Prec}+\mathrm{Cov}}, & \mathrm{Prec}+\mathrm{Cov}>0,\\
0, & \text{otherwise},
\end{cases} \\
\mathrm{False\ label\ rate} &=
\begin{cases}
\frac{L}{M}, & M>0,\\
0, & M=0,
\end{cases}
&
\mathrm{Efficiency} &=
\begin{cases}
\frac{C}{W}, & W>0,\\
0, & W=0.
\end{cases}
\end{align}
\end{subequations}
where $\mathrm{Prec}$ and $\mathrm{Cov}$ denote precision and coverage, respectively.

Accuracy is the primary metric. Coverage and precision separate failure to elicit information from failure to interpret information once stated. False label rate captures incorrect extractions among extracted fields, and efficiency measures the correct fields recovered per cumulative word.

\begin{table}[h]
\raggedright
\small
\caption{Pooled benchmark results across all 300 sessions, reported as mean $\pm$ standard deviation. Up arrows indicate higher-is-better metrics; down arrows indicate lower-is-better metrics. Best values in each column are bolded. Standard deviations are computed across session-level metric values. The reported mean $\pm$ standard deviation is descriptive only, so the implied range may exceed $[0,100]\%$ even though individual accuracies are bounded.}
\label{tab:main_results}

\vspace{0.3em}
\begin{tabular}{@{}lcccc@{}}
\toprule
\textbf{Strategy} & \textbf{Accuracy $\uparrow$} & \textbf{Coverage $\uparrow$} & \textbf{Precision $\uparrow$} & \textbf{F1 score $\uparrow$} \\
\midrule
Random     & 51.7\% $\pm$ 19.6\% & 52.4\% $\pm$ 19.5\% & 98.4\% $\pm$ 6.9\% & 66.1\% $\pm$ 18.8\% \\
Form-based & 84.8\% $\pm$ 15.8\% & 85.5\% $\pm$ 15.8\% & 99.2\% $\pm$ 2.8\% & 90.9\% $\pm$ 11.5\% \\
LLM-guided & \textbf{95.4\% $\pm$ 8.1\%} & \textbf{96.0\% $\pm$ 7.8\%} & \textbf{99.4\% $\pm$ 2.5\%} & \textbf{97.5\% $\pm$ 4.6\%} \\
\bottomrule
\end{tabular}

\vspace{0.6em}

\begin{tabular}{@{}lcccc@{}}
\toprule
\textbf{Strategy} & \textbf{False label rate $\downarrow$} & \textbf{Efficiency $\uparrow$} & \textbf{Mean turns $\downarrow$} & \textbf{Mean words $\downarrow$} \\
\midrule
Random     & 1.6\% $\pm$ 6.9\% & 0.0090 $\pm$ 0.0040 & 20.0 $\pm$ 0.0 & 645.2 $\pm$ 318.7 \\
Form-based & 0.8\% $\pm$ 2.8\% & 0.0197 $\pm$ 0.0067 & 20.0 $\pm$ 0.0 & \textbf{489.7 $\pm$ 213.3} \\
LLM-guided & \textbf{0.6\% $\pm$ 2.5\%} & \textbf{0.0207 $\pm$ 0.0097} & \textbf{18.3 $\pm$ 3.1} & 589.8 $\pm$ 309.9 \\
\bottomrule
\end{tabular}
\end{table}

\section{Experiments}
Each interview follows a fixed loop: select a question, generate the patient response, update the transcript, and update the in-loop recovery state. Sessions stop when the interview reaches 20 turns or when all target fields have been recovered.

All experiments used GPT-4o-based components. The patient agent used temperature 0.7 with a maximum of 1024 output tokens. The LLM-guided selector used temperature 0.3. The judge used temperature 0.0 with a maximum of 512 output tokens. The benchmark uses a full factorial design with 3 policies, 4 evaluation patients, 5 behavioral conditions, and 5 repeated runs per cell, yielding 300 interview sessions in total; repeated runs capture stochasticity in the random and LLM-guided policies, while the form-based baseline is deterministic given its fixed ordering.

Table~\ref{tab:main_results} reports pooled performance across all 300 sessions. The overall strategy hierarchy is consistent across the full benchmark: random performs worst, form-based improves substantially over random, and LLM-guided performs best. Random reaches 51.7\%\(\pm\)19.6\% mean accuracy, while form-based reaches 84.8\%\(\pm\)15.8\% and LLM-guided reaches 95.4\%\(\pm\)8.1\%. The 33.1 percentage point gap between random and form-based indicates that clinical structure accounts for most of the gain, while the additional 10.6 point gap between form-based and LLM-guided reflects the value of adapting to patient responses in real time. These pooled results suggest that the main difference across strategies lies in whether the relevant topics are reached within the interaction budget rather than in how accurately extracted information is labeled once disclosed.

\begin{figure}[t]
\centering
\includegraphics[width=0.8\linewidth]{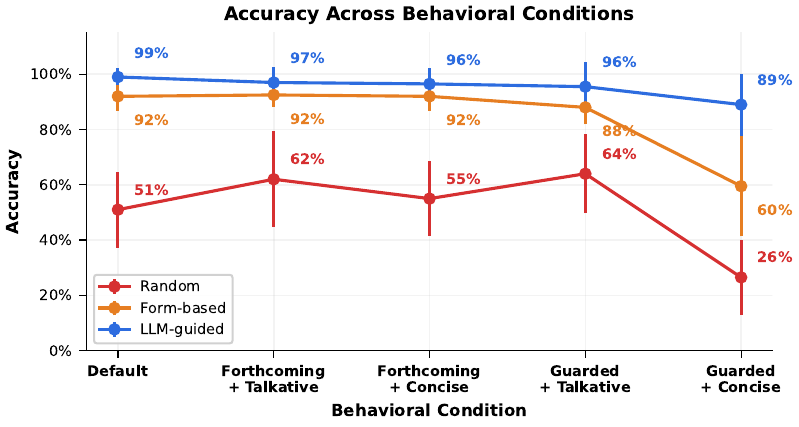}
\caption{Mean accuracy across the five behavioral conditions. The form-based and LLM-guided strategies remain strong through Guarded + Talkative, but the form-based and random policies degrade sharply under \emph{Guarded + Concise}, while the LLM-guided policy retains 89\% accuracy.}
\label{fig:degradation}
\end{figure}

Figure~\ref{fig:degradation} summarizes performance across the five behavioral conditions. Across the five explicit behavioral conditions, the same overall ordering holds: random performs worst, form-based improves substantially over random, and LLM-guided performs best. Under Default, Forthcoming + Talkative, Forthcoming + Concise, and Guarded + Talkative, the form-based and LLM-guided strategies remain strong, while random remains consistently lower. Under Guarded + Concise, however, performance drops sharply for the random and form-based strategies, while the LLM-guided policy retains 89\% accuracy.

Across the first four conditions, the clinically ordered form remains relatively close to the adaptive LLM policy. Under Guarded + Concise, however, the LLM--form gap expands to 29 percentage points, showing that adaptation matters most when patients do not quickly reveal target fields.

\begin{figure}[t]
\centering
\includegraphics[width=0.8\linewidth]{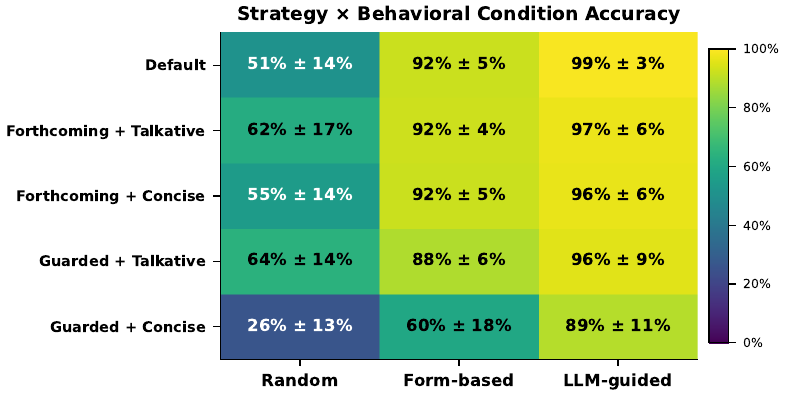}
\caption{Strategy accuracy across the five behavioral conditions. The LLM-guided policy performs best in every condition, while the form-based baseline drops sharply under \emph{Guarded + Concise}.}
\label{fig:heatmap}
\end{figure}

\paragraph{Performance by Behavioral Condition:} Figure~\ref{fig:heatmap} shows strategy accuracy under all five behavioral conditions. First, LLM-guided selection remains the strongest method in every condition, ranging from 89.0\%\(\pm\)11.2\% under guarded-concise to 99.0\%\(\pm\)3.1\% under default conditions. Second, the form-based baseline is highly competitive in non-guarded settings, staying near 92\%--93\% across default and forthcoming conditions, but drops sharply to 59.5\%\(\pm\)18.2\% under guarded-concise. Random remains substantially lower throughout, ranging from 26.5\%\(\pm\)13.5\% to 64.0\%\(\pm\)14.3\%. The heatmap also shows that the benchmark is not simply measuring overall patient difficulty. Talkativeness partially offsets guardedness by increasing incidental disclosure, whereas guarded-concise exposes the limitation of fixed one-question-per-topic interviewing.

\begin{figure}[t]
\centering
\includegraphics[width=0.8\linewidth]{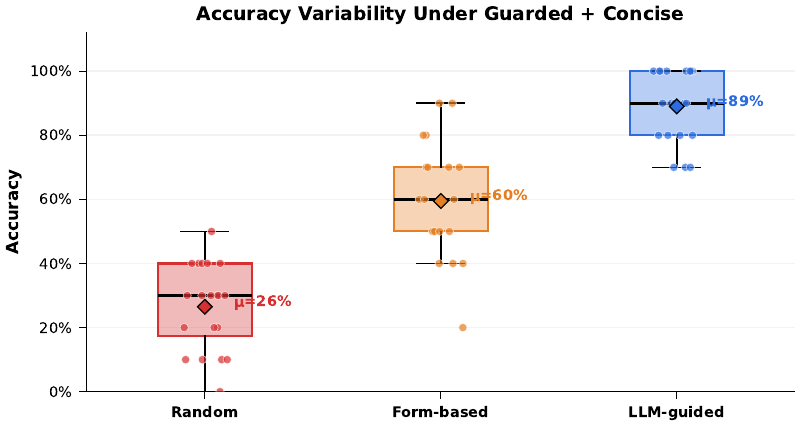}
\caption{Per-session accuracy under \emph{Guarded + Concise}. The LLM-guided policy is both more accurate and less variable than the form-based and random baselines under this condition.}
\label{fig:hard_variability}
\end{figure}

\paragraph{Hard-Condition Variability:} Figure~\ref{fig:hard_variability} focuses on the \emph{Guarded + Concise} condition,  which is the hardest setting in the benchmark. In this regime, random centers around 26.5\%\(\pm\)13.5\% accuracy, form-based around 59.5\%\(\pm\)18.2\%, and LLM-guided around 89.0\%\(\pm\)11.2\%. This is also the setting in which the fixed form collapses most clearly, while the LLM policy remains substantially more robust -- the form strategy asks one question per topic and then moves on, whereas the LLM policy can respond to deflections by re-asking or switching to a better-targeted follow-up.

\begin{figure}[t]
\centering
\includegraphics[width=0.8\linewidth]{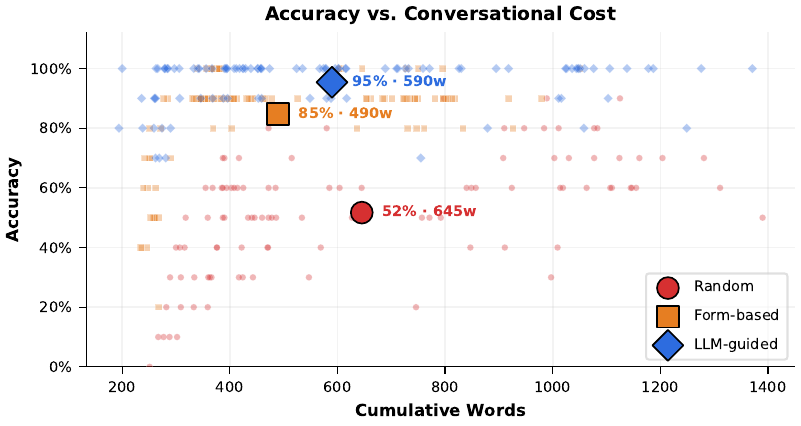}
\caption{Accuracy versus conversational cost. The LLM-guided policy achieves the highest accuracy, the form-based policy is most word-efficient on average, and the random policy uses the most words while recovering the least information.}
\label{fig:efficiency}
\end{figure}

\paragraph{Efficiency Analysis:} Figure~\ref{fig:efficiency} compares transcript accuracy against cumulative word count across all 300 sessions. LLM-guided selection achieves the highest overall accuracy while remaining efficient in conversational cost, with the pooled centroid in Figure~\ref{fig:efficiency} lying near 95\% accuracy at roughly 590 words. Form-based selection is also efficient, centering near 85\% at roughly 490 words, while random reaches only about 52\% at roughly 645 words. All three strategies operate under the same 20-turn budget, so the comparison primarily reflects what is asked rather than the available number of questions.

These results show that the strongest strategy is not merely eliciting longer transcripts. Instead, it recovers more clinically relevant information per unit of conversational effort.

\paragraph{Patient-Level Heterogeneity:} Table~\ref{tab:patient_results} summarizes pooled accuracy by the synthetic patient. For example, Eval004, a PTSD veteran profile with 8/10 "yes" fields in the binary evaluation, is the easiest case for form-based and LLM-guided selection, reaching 92.8\% and 97.2\% respectively. Eval001 is the hardest case for the form-based baseline at 80.8\%, while Eval003 is the hardest case for random selection at 45.2\%. More broadly, patients with more "yes" fields tend to be easier for random questioning because off-target questions can still trigger incidental disclosure, whereas sparse mostly-negative profiles depend more strongly on explicit confirmation.

\begin{table}[t]
\caption{Pooled accuracy by patient across all behavioral conditions, reported as mean $\pm$ standard deviation over 25 sessions per patient-strategy cell (5 conditions $\times$ 5 runs). Best value in each row is bolded.}
\label{tab:patient_results}
\centering
\small
\begin{tabular}{lccc}
\toprule
\textbf{Patient} & \textbf{Random} & \textbf{Form-based} & \textbf{LLM-guided} \\
\midrule
Eval001 (MDD+GAD) & 47.2\% $\pm$ 19.0\% & 80.8\% $\pm$ 16.1\% & \textbf{94.4\% $\pm$ 8.2\%} \\
Eval002 (MDD+SI)  & 55.6\% $\pm$ 21.8\% & 82.4\% $\pm$ 14.5\% & \textbf{96.0\% $\pm$ 8.2\%} \\
Eval003 (GAD)     & 45.2\% $\pm$ 16.6\% & 83.2\% $\pm$ 17.3\% & \textbf{94.0\% $\pm$ 9.6\%} \\
Eval004 (PTSD)    & 58.8\% $\pm$ 18.6\% & 92.8\% $\pm$ 13.1\% & \textbf{97.2\% $\pm$ 6.1\%} \\
\bottomrule
\end{tabular}
\end{table}

\subsection{Precision and Coverage}
Precision remains uniformly high across all three strategies, ranging from 98.4\% for random to 99.4\% for LLM-guided. This narrow range contrasts with the much larger gaps in coverage and accuracy. The practical implication is that the benchmark is not bottlenecked by transcript interpretation once information has been stated. Instead, the main challenge is whether the interview strategy reaches the right topics and elicits the relevant information within the available interaction budget.

False label rates are low across all three strategies and highest for random questioning, consistent with the tendency of random policies to produce more tangential transcripts in which the judge must infer field values from indirect or ambiguous context.

\section{Discussion}

This benchmark yields three main findings. First, question selection matters substantially in conversational psychiatric intake. Relative to random questioning, the clinically ordered form-based baseline improves performance by a large margin, showing that much of the gain comes from imposing appropriate clinical structure on the interview. Second, adaptation still matters beyond structure. Although the form-based baseline performs strongly across most behavioral conditions, the LLM-guided policy achieves the best overall recovery, and its advantage grows sharply under guarded-concise patients. Third, the benchmark suggests that the main bottleneck in this setting is not transcript interpretation once information is disclosed, but whether the interview reaches the binary evaluation topics within a limited interaction budget.

Of note, random questioning performs poorly for a structural reason. The question bank is clinically grounded and intentionally broader than the 10 benchmark targets, so many candidate questions are plausible in intake while still being low-yield for the measured recovery objective. A policy that samples broadly without prioritization spends a substantial fraction of its turn budget on content that does not advance recovery. The large gap between random and form-based performance shows that clinically meaningful ordering alone resolves much of this inefficiency.

In addition, a fixed form approach is highly competitive when patients are forthcoming, because a clinically ordered sequence already reaches most high-yield domains within the available interaction budget. When patient behavior is less amenable to field recovery, however, a fixed one-question-per-topic strategy becomes brittle. When patients defer, partially answer, or provide minimal responses, the interviewer must decide whether to re-ask, reframe, or shift to a better-targeted follow-up. The LLM-guided policy performs better in these settings because it conditions on the evolving dialogue state and can respond to local topic shifts and partial disclosure. In that sense, this benchmark clarifies that structure explains most of the gain over random, while adaptation explains most of the remaining gain under difficult conversational conditions.

Across all three strategies, precision remains uniformly high but coverage varies substantially. This pattern suggests that the benchmark is not primarily limited by post-hoc extraction once the relevant information appears in the transcript. Instead, the more consequential failure mode is upstream: whether the questioning policy elicits the right information at all. For interactive clinical machine learning, this implies that stronger language understanding alone is not sufficient if the system does not ask about the right topics at the right time.

Finally, these findings support that conversational clinical systems should not be evaluated by average task performance alone under default conditions on patient behavioral condition. In this setting, average recovery, degradation under behavior that is less amenable to field recovery, conversational efficiency, and robustness to behavioral variation all matter. The benchmark also does not show that an automated system should replace a clinician during psychiatric intake. Rather, it provides a controlled way to compare questioning policies and to study how clinical structure and adaptive follow-up affect information recovery under realistic interaction constraints, but still subject to the limitations of synthetic patient representation. 

\subsection{Limitations and Future Work}
 A key limitation is that patients are simulated rather than real, so the results should be interpreted as benchmark results rather than direct estimates of clinical performance. In addition, while the question bank is grounded in real clinician-authored intake materials, it reflects a single institutional workflow and a benchmark abstraction of only 10 binary targets, rather than more subjective assessments of affect or chief complaint, which were more difficult to set up an evaluation framework for. Although the benchmark includes 300 total interview sessions, these are generated from only four evaluation patients, corresponding to 100 patient–condition–run combinations that are then evaluated across the three questioning strategies. The patient set was selected to maximize heterogeneity rather than to estimate population-level performance. In addition, the same LLM family is used across multiple system components, which may introduce shared biases. Finally, the benchmark does not yet include a human clinician baseline, and although LLM-based judges are increasingly used as scalable evaluation tools, they can introduce systematic biases. Thus the present findings should be interpreted as \textit{comparative} results between these strategies \citep{zheng2023judging} rather than some ground truth. 

The results support expanding the framework in the ways described above, and extending it in new directions. Primarily, we would like to compare the adaptive questioning from a question bank (this study's task) to a fully unconstrained human evaluation with these patient agents. This would support a longer-term direction to build a clinician-facing training and evaluation platform centered on reliable synthetic patient interactions. This next stage will require continued collaboration between machine learning researchers and clinical stakeholders to ensure that the evaluated task remains meaningful \citep{saleh2020clinical,wiens2019donoharm}.

\section{Conclusion}

This paper introduced a benchmark for question selection in conversational psychiatric intake. The benchmark is grounded in real clinician-authored intake questions, evaluated against known recovery targets, and stress-tested under different patient behavioral conditions. Across the benchmark, random questioning performed worst, a clinically ordered fixed form substantially improved recovery, and an LLM-guided policy achieved the best overall performance. The results indicated that clinical structure explains much of the gain over random questioning, while adaptation explains much of the remaining gain when patient behavior is less amenable to field recovery. These findings suggest that psychiatric intake can be meaningfully framed as a budgeted information-recovery problem under a limited interaction budget, and under that framing, this benchmark provides a controlled foundation for future clinician-facing training, assessment, and decision-support systems.

\newpage

\bibliography{references}

\newpage

\appendix

\section{Construction of the Form-Based Baseline}

A clinically ordered sequence was constructed offline because the native bank order reflected source-sheet parsing rather than a reproducible psychiatric assessment protocol. Under the 20-turn budget, using the raw bank order would leave several high-value targets unreached and would therefore produce an artificially weak fixed-form comparator. The final ordering was built through four layers of evidence-based decisions, summarized in Table~\ref{tab:form_ordering}.

\begin{table}[t]
\centering
\caption{Evidence layers used to construct the form-based baseline.}
\label{tab:form_ordering}
\small
\begin{tabular}{p{2.6cm}p{4.8cm}p{4.8cm}}
\toprule
\textbf{Design layer} & \textbf{Decision} & \textbf{Primary justification} \\
\midrule
Domain sequence & Presenting problem $\rightarrow$ risk $\rightarrow$ psychiatric history $\rightarrow$ medical history $\rightarrow$ social/family context & APA psychiatric evaluation structure; modular interview design in SCID and MINI \citep{apa2006evaluation,first2016scid,sheehan1998mini} \\
\addlinespace
Safety elevation & Move suicide-risk screening to the start of the interview & Early suicide-risk assessment \citep{posner2011cssrs,jointcommission2018r3suicide} \\
\addlinespace
Budget allocation & Per-domain turn caps under the 20-turn budget & Structured interviews distribute attention across clinically important modules \\
\addlinespace
Within-domain ordering & Prioritize high-yield screening items; collapse redundant variants; prefer binary questions aligned to benchmark targets & Clinical screening logic and target coverage under limited turns \\
\bottomrule
\end{tabular}
\end{table}

The source intake forms served as the clinical origin of the question bank and as external validation of domain coverage, but not as the primary justification for the final order. These forms were designed for written pre-visit completion rather than conversational assessment under a fixed turn budget, and no single published instrument directly targets all 10 benchmark fields within a 20-turn conversational budget.

\end{document}